\documentclass[conference]{ieeeconf}
\usepackage{times}
\usepackage[bookmarks=true]{hyperref}
\usepackage{booktabs} 
\usepackage[ruled,vlined,titlenotnumbered,linesnumbered]{algorithm2e} 

\usepackage{graphicx,caption,subcaption}
\usepackage{amsmath,amssymb,stmaryrd,mathtools}
\usepackage[ruled,vlined,titlenotnumbered,linesnumbered]{algorithm2e} 
\usepackage[noend]{algpseudocode}
\usepackage{acronym}
\usepackage{comment}
\usepackage{color}
\usepackage{units}
\usepackage{color, colortbl}
\usepackage{adjustbox}
\usepackage{array,multirow}

\definecolor{darkGray}{gray}{0.7}
\definecolor{lightGray}{gray}{0.9}

\definecolor{turquoise}{RGB}{114, 189, 204}
\definecolor{lightTurquoise}{RGB}{195, 237, 244}

\definecolor{purple}{RGB}{158, 118, 191}
\definecolor{lightPurple}{RGB}{232, 204, 255}

\usepackage[capitalise]{cleveref}

\DeclareCaptionLabelSeparator{periodspace}{.\quad}
\crefname{equation}{}{} 
\crefname{section}{Sec.}{Sec.}

\newcommand{\tvar}{t}
\newcommand{\R}{\mathbb{R}}
\newcommand{\ctrl}{u}
\newcommand{\dstb}{d}
\newcommand{\cfunc}{u(\cdot)}
\newcommand{\dfunc}{d(\cdot)}
\newcommand{\cset}{\mathcal{U}}
\newcommand{\cfset}{\mathbb{U}}
\newcommand{\dset}{\mathcal{D}}
\newcommand{\dfset}{\mathbb{D}}
\newcommand{\state}{x}
\newcommand{\traj}{\xi} 
\newcommand{\dyn}{f} 
\newcommand{\targetfunc}{l}
\newcommand{\targetset}{\mathcal{L}}
\newcommand{\costfunctional}{J}

\newcommand{\vfunc}{V}
\newcommand{\brs}{\mathcal{V}} 

\newcommand{\ham}{H}

\newcommand{\pos}{p} 
\newcommand{\npos}{h} 
\newcommand{\tdummy}{\tau}
\newcommand{\tdummyy}{s}
\newcommand{\trajstandard}{\traj_{\state, \tdummy}^{\ctrl,\dstb}}

\newcommand{\env}{\mathcal{E}}
\newcommand{\planner}{\Pi}
\newcommand{\sense}{\mathcal{S}}
\newcommand{\initSafe}{\mathcal{X}_{\text{init}}}
\newcommand{\brsSafe}{\mathcal{W}}
\newcommand{\horizon}{T}

\newcommand{\freeRegion}{\mathcal{F}}
\newcommand{\map}{\mathcal{M}}
\newcommand{\queue}{\mathcal{Q}}
\newcommand{\neighbor}{\mathcal{N}}
\newcommand{\timestep}{\Delta{T}}

\newcommand{\reals}{\mathbb{R}}

\newtheorem{lemma}{Lemma}

\IEEEoverridecommandlockouts

\begin{document}
\title{\LARGE \bf An Efficient Reachability-Based Framework for Provably Safe Autonomous Navigation in Unknown Environments} 
\author{Andrea Bajcsy*, Somil Bansal*, Eli Bronstein, Varun Tolani, Claire J. Tomlin
\thanks{*Equal contribution. 
All authors are with the Department of Electrical Engineering and Computer Sciences, University of California, Berkeley.
}
}
\maketitle

\begin{abstract}
Real-world autonomous vehicles often operate in \textit{a priori} unknown environments. Since most of these systems are safety-critical, it is important to ensure they operate safely in the face of environment uncertainty, such as unseen obstacles. Current safety analysis tools enable autonomous systems to reason about safety given full information about the state of the environment \textit{a priori}. However, these tools do not scale well to  scenarios where the environment is being sensed in real time, such as during navigation tasks. In this work, we propose a novel, real-time safety analysis method based on Hamilton-Jacobi reachability that provides strong safety guarantees despite environment uncertainty. Our safety method is planner-agnostic and provides guarantees for a variety of mapping sensors. We demonstrate our approach in simulation and in hardware to provide safety guarantees around a state-of-the-art vision-based, learning-based planner.  
Videos of our approach and experiments are available on the project website\footnote{Project website: \href{https://abajcsy.github.io/safe\_navigation/}{\textcolor{blue}{https://abajcsy.github.io/safe\_navigation/}}}.
\end{abstract}

\section{Introduction} \label{sec:intro}
Autonomous vehicles operating in the real world must navigate through \textit{a priori} unknown environments using on-board, limited-range sensors.
As a vehicle makes progress towards a goal and receives new sensor information about the environment, rigorous safety analysis is critical to ensure that the system's behavior does not lead to dangerous collisions. In order to provide such safety guarantees for real vehicles, any analysis should take into account multiple sources of uncertainty, such as modelling error, external disturbances, and unknown parts of the environment. 

A variety of mechanisms have been proposed to ensure robustness to modeling error and external disturbances \cite{majumdar2017funnel, herbert2017fastrack, singh2017robust}.
Additionally, safety guarantees for systems using limited-range sensors in unknown environments have been the subject of recent investigation \cite{kousik2018bridging, Lahijanian16tro, Sarid12guaranteeinghighlevel, fridovich2018safe}.
Although interesting results have emerged from these studies, the safety guarantees are provided by imposing specific assumptions on the sensor and/or the planner that are rather restrictive for a variety of real-world autonomous systems and sensors used for navigational purposes.
In contrast, rather than proposing a specific planning and sensing paradigm that guarantees safety, we aim to design a safety framework that is compatible with a broad class of sensors, planners, and dynamics.

There are two main challenges with providing such a framework. The first challenge relates to ensuring safety with respect to unknown parts of the environment and external disturbances while minimally interfering with goal-driven behavior. 
Second, real-time safety assurances need to be provided as new environment information is acquired, which requires approximations that are both computationally efficient and not overly conservative. Moreover, this safety analysis should be applicable to a wide variety of real-world sensors, planners, and vehicles.
\begin{figure} [t!]
    \centering
    \includegraphics[width=\columnwidth]{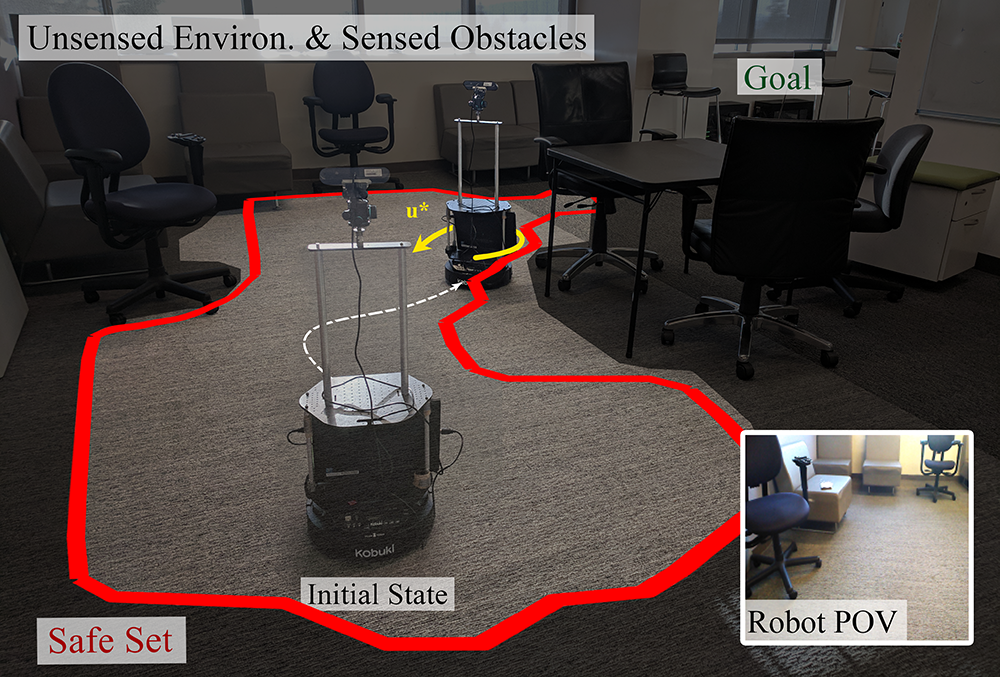}
    \caption{\textbf{Overview:} We consider the problem of safe navigation from an initial state to a goal state in an \textit{a priori} unknown environment. 
    Our approach treats the unsensed environment as an obstacle, and uses a HJ reachability framework to compute a safe controller for the vehicle, which is updated in real-time as the vehicle explores the environment. 
    We show an application of our approach on a Turtlebot using a vision-based planner. When the robot is at risk of colliding, the safe controller ($\ctrl^{*}$) keep the system safe.}
    \label{fig:hardware_front_fig}
\end{figure}

In this paper, we propose a safety framework that can overcome these challenges for autonomous vehicles operating in \textit{a priori} unknown static environments under the assumption that the sensors work perfectly within their ranges.
Erroneous and noisy sensors can make safety analysis significantly more challenging and we defer this to future work.
Our framework is based on Hamilton Jacobi (HJ) reachability analysis \cite{Margellos11, mitchell2005time}, a verification method for guaranteeing safety and performance for dynamical systems with general nonlinear dynamics and disturbances. This approach provides not only the set of states from which the dynamical system can always satisfy state constraints (i.e. remain collision-free), but also provides an optimal controller that guarantees the system will never violate the state constraints. 
In particular, we treat the unknown environment at any given time as an obstacle and use HJ reachability to compute the \textit{backward reachable set (BRS)}, i.e. the set of states from which the vehicle can enter the unknown and potentially unsafe part of the environment, despite the best control effort. The complement of the BRS therefore represents the safe set for the vehicle.
With this computation, we also obtain the corresponding least restrictive safety controller, which does not interfere with the planner unless the safety of the vehicle is at risk. 
Use of HJ reachability analysis in our framework thus allows us to overcome the first challenge---our framework can be applied to general nonlinear vehicles, sensors, and planners. 

In general, due to the computationally expensive nature of HJ computations, this approach has not been leveraged in settings where the environment is not known beforehand and rather is sensed at run-time.
To overcome this challenge, we propose a novel, real-time algorithm to compute the BRS.
Our algorithm only locally updates the BRS in light of new environment information, which significantly alleviates the computational burden of HJ reachability while still maintaining the safety guarantees at all times. 
To summarize, our key contributions are:
\begin{enumerate}
    \item a provably-safe framework for navigation in static unknown environments that is applicable to a broad class of sensors, planners, and dynamics,
    \item an algorithm for online safe set updates from new sensor measurements as the robot navigates,
    \item demonstration of our approach on different sensors and planners on a vehicle with nonlinear dynamics in the presence of external disturbances,
    \item a hardware demonstration of our approach to provide safety around a state-of-the-art learning-based planner which uses only monocular RGB images for planning.
\end{enumerate} 

\section{Related Work} \label{sec:related_work}
An extensive body of research deals with motion planning and safe exploration for robots in unknown environments, some of which focuses on safety guarantees despite modeling error and external disturbances. We cannot hope to summarize all these works here, but we attempt to discuss several of the most closely related approaches.

\subsection{Safe motion planning}
Methods that ensure safety despite modeling error and disturbances are largely motivated by the trade-off between safety and efficiency during real-time planning. 
A popular approach is to perform \textit{offline} computations that quantify disturbances and modeling error which can be used \textit{online} to determine collision-free trajectories \cite{majumdar2017funnel, herbert2017fastrack, singh2017robust}. 
Alternatively, \cite{agrawal2017discrete, ames2017control} use control barrier functions to design provably stable controllers while satisfying given state-space constraints.
However, these methods assume that a recursively feasible collision-free path can be obtained despite the unknown environment, which may not be possible in real-world environments.
Several works address this problem for single-agent scenarios within a model predictive control framework \cite{richards2003model, rosolia2018learning}, as well as for multiple vehicles using sequential trajectory planning \cite{schouwenaars2004decentralized, bansal2017safe}. However, these works assume \textit{a priori} knowledge of all obstacles, whereas our framework guarantees safety in an \textit{a priori} unknown environments for potentially high-order nonlinear dynamics.

Ensuring safety with respect to both modeling error and limited sensing horizons have been studied using sum-of-squares \cite{kousik2018bridging}, linear temporal logic \cite{Lahijanian16tro}, reactive synthesis approaches \cite{Sarid12guaranteeinghighlevel}, graph-based kinodynamic planner \cite{fridovich2018safe} among others.
These works typically impose restrictions on sensors or planners to ensure safety with respect to the unknown environment. 
In contrast, the proposed framework is sensor and planner agnostic, provided that the sensor can accurately identify the obstacles within it's sensing region.

\subsection{Safe exploration}
The problem of finding feasible trajectories to a specified goal in an unknown environment has also been studied in the robotic exploration literature for simplified kinematic motion models using frontier-exploration methods \cite{yoder2016autonomous} and D* \cite{koenig2005fast}. 
Other works include sampling-based motion planners for drift-less dynamics \cite{bekris2007greedy} and dynamic exploration methods for vehicles with a finite stopping time \cite{janson2018safe}. 
Robotic exploration has been also studied within the context of fully and partially observable Markov decision processes \cite{richter2018bayesian, moldovan2012safe} and reinforcement learning \cite{achiam2017constrained, kahn2017uncertainty} 
to reduce collision probabilities; however, no theoretical safety guarantees are typically provided.

Safe exploration has also been studied in terms of Lyapunov stability \cite{berkenkamp2017safe, chow2018lyapunov}.
Even though stability is often desirable, it is insufficient to guarantee collision avoidance.
In contrast, our formulation uses a stronger definition of safety, and is more in line with \cite{fraichard2004inevitable, fisac2018general}, which characterize safety using reachable sets.

\section{Problem Statement} \label{sec:formulation}
In this work, we study the problem of autonomous navigation in \textit{a-priori unknown static} environments.
Consider a stable, deterministic, nonlinear dynamical model of the vehicle
\begin{equation} \label{eqn:dynamics}
\dot{\state} = \dyn(\state, \ctrl, \dstb),
\end{equation}
where $\state \in \reals^{n}$, $\ctrl \in \cset$, and $\dstb \in \dset$ represent the state, the control, and the disturbance experienced by the vehicle. Here, $\dstb$ can include the effect of both the external disturbances or dynamics mismatch. 
For convenience, we partition the state $\state$ into the position component $\pos \in \reals^{n_\pos}$ and the non-position component $\npos \in \reals^{n - n_\pos}$: $\state = (\pos, \npos)$.
We assume that the flow field $\dyn: \reals^{n}\times\cset\times\dset \rightarrow \reals^{n}$ is uniformly continuous in time, and Lipschitz continuous in $\state$ for fixed $\ctrl$ and $\dstb$. 
With this assumption, given $\ctrl(\cdot) \in \cfset, \dstb(\cdot) \in \dfset$, there exists a unique trajectory solving \eqref{eqn:dynamics}~\cite{EarlA.Coddington1955}. 
We also assume that the vehicle state $\state$ can be accurately sensed at all times. 

Let $\state_0$ and $\state^{*}$ denote the start and the goal state of the vehicle. 
The vehicle aims to navigate from $\state_0$ to $\state^{*}$ in an \textit{a priori} unknown environment, $\env$, whose map or topology is not available to the robot. 
At any time $\tvar$ and state $\state(\tvar)$, the vehicle has a planner $\planner(\state(\tvar), \state^{*}, \env)$, which outputs the control command $\ctrl(\tvar)$ to be applied at time $\tvar$.
The vehicle also has a sensor which at any given time exposes a region of the state space $\sense_\tvar \subset \reals^{n}$, and provides a conservative estimate of the occupancy within $\sense_\tvar$. 
For example, if the vehicle has a camera sensor, $\sense_\tvar$ would be a triangular region (prismatic in 3D) representing the field-of-view of the camera.
We assume perfect perception within this limited sensor range.
Dealing with erroneous perception, sensor noise, and dynamic environments are problems in their own right, and we defer them to future work.
Finally, we assume that there is a known initial obstacle-free region around $\state_0$ given by $\initSafe \subset \reals^{n}$; e.g. this is the case when the vehicle is starting at rest and its initial state is collision-free.

Given $\state_0$, $\state^{*}$, $\initSafe$, the planner $\Pi$, and the sensor $\sense$, the goal of this paper is to design a least restrictive control mechanism to navigate the vehicle to the goal state while remaining safe, which means avoiding obstacles at all times.
Since the environment $\env$ is unknown, the safety needs to be ensured given the partial observations of the environment obtained through the sensor, which in general is challenging. 
We use the HJ reachability-based framework to ensure safety despite only partial knowledge of the environment.

\section{Running Example} \label{sec:running_example}
To illustrate our approach and provide intuition behind the proposed framework, we introduce a simple running example: a 3-dimensional Dubins' car system with disturbances added to the velocity. The dynamics of the system are given by:
\begin{equation} \label{eqn:3D_dubins_dyn}
\begin{aligned}
\dot{p}_x = v\cos\phi + d_x,\quad \dot{p}_y = v\sin\phi + d_y,\quad \dot{\phi} = \omega\,, \\
\underline{v} \le v \le \bar{v},\quad |\omega| \le \bar{\omega},\quad |d_{x}|, |d_{y}| \le d_{r}
\end{aligned}
\end{equation}
where $\state := (p_x, p_y, \phi)$ is the state, $\pos = (p_x, p_y)$ is the position, $\phi$ is the heading, and $d = (d_{x}, d_{y})$ is the disturbance experienced by the vehicle. 
The control of the vehicle is $\ctrl := (v, \omega)$, where $v$ is the speed and $\omega$ is the turn rate.
Both controls have a lower and upper bound, which for this example are chosen to be $\underline{v} = 0.1m/s$, $\bar{v} = 1m/s$, $\bar{\omega}=1rad/s$.
The disturbance bound is chosen as $d_{r} = 0.1m/s$.

\begin{figure} [h!]
    \centering
    \begin{subfigure}[b]{0.55\columnwidth}
    \centering
    \includegraphics[width=1.0\columnwidth]{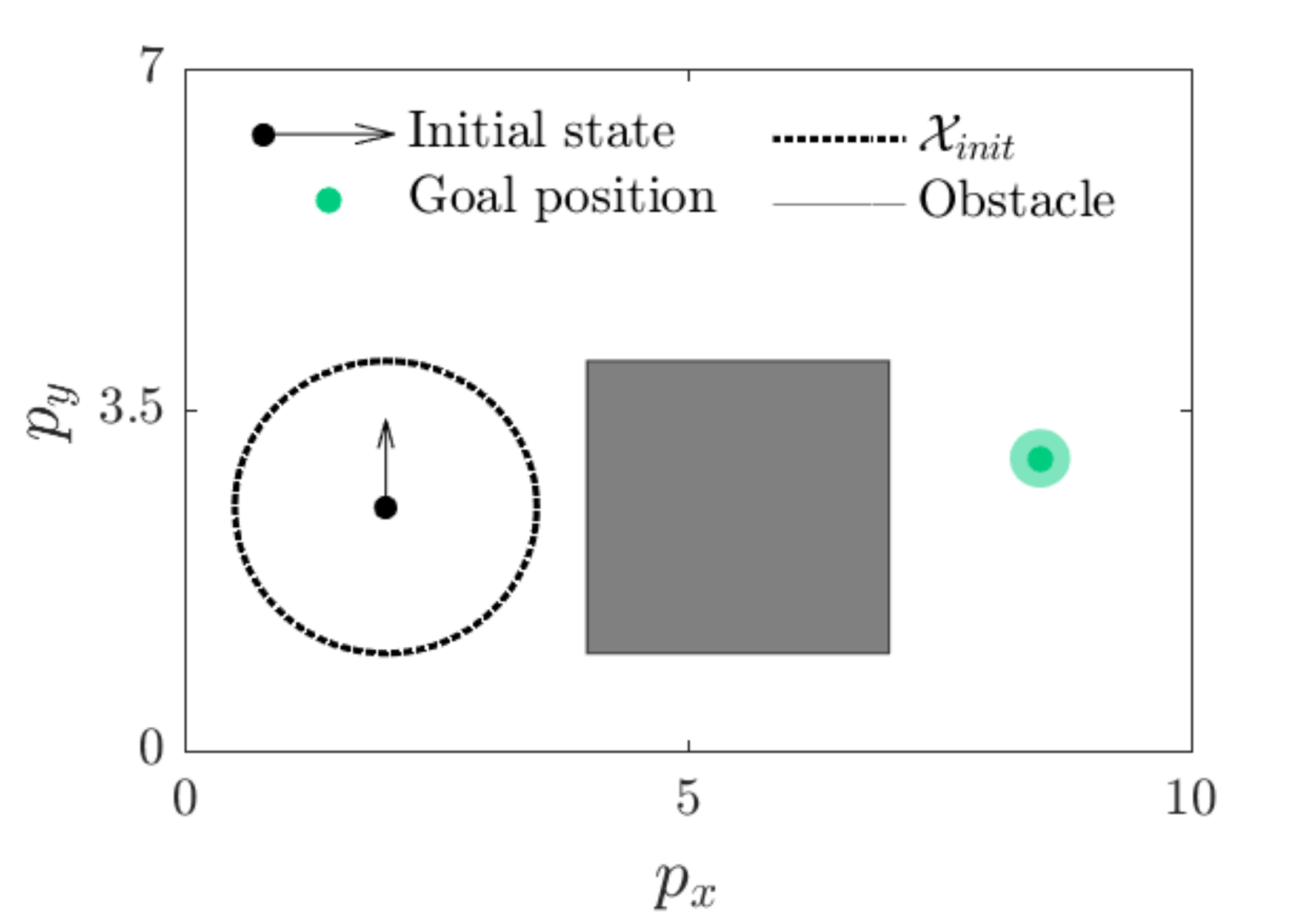}
      \label{fig:init_setup1}
    \end{subfigure}%
    \centering
    \begin{subfigure}[b]{0.45\columnwidth}
    \centering
    \begin{subfigure}[b]{\columnwidth}
    \centering
    \includegraphics[width=0.5\columnwidth]{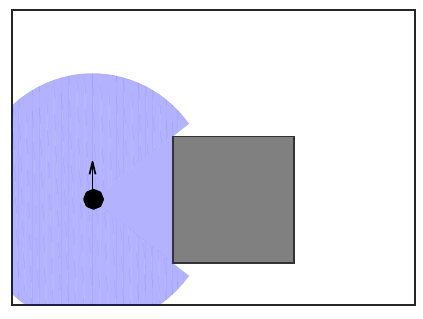}
    \subcaption{LiDAR sensing region}
      \label{fig:init_setup2}
    \end{subfigure}%
    \centering
    
    \begin{subfigure}[b]{\columnwidth}
    \centering
    \includegraphics[width=0.5\columnwidth]{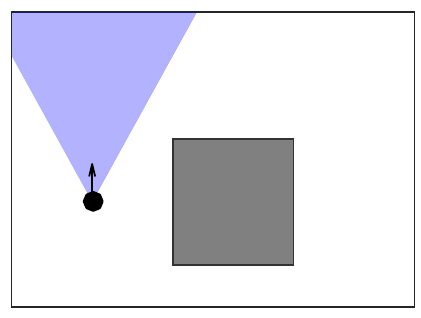}
    \subcaption{Camera sensing region}
      \label{fig:init_setup3}
    \end{subfigure}%
    \end{subfigure}%
    \caption{The initial setup for the running example. 
    The goal is to safely reach the goal (center of the green area) from the initial position (black marker) in the presence of an unknown obstacle (the grey square).
    We also show the initial sensing region for the LiDAR and camera sensors.} 
    \label{fig:initial_setup_running_example}
\end{figure}

The environment setup for is shown in Figure~\ref{fig:initial_setup_running_example}.
The vehicle start and the goal state are given by $\state_0 = [2, 2.5, \pi/2]$ (shown in black) and $\state^* = [8.5, 3, -\pi/2]$ (the center of the green area).
The goal is to reach within $0.3m$ of $\state^*$ (the light green area).
However, there is an obstacle in the environment which is not known to the vehicle beforehand (shown in grey). 
At the beginning of the running example navigation task, we assume that there is no obstacle within $1.5m$ of $\state_0$, and obtain the initial obstacle-free region $\initSafe := \{\state: \|\pos - \pos_0\| \le 1.5\}$ (the area inside the dashed black line).

To demonstrate the sensor-agnostic nature of our approach, we simulate the Dubins' car with two different sensors: a LiDAR and a camera.
For a LiDAR, the sensing region $\sense_\tvar$ is given by a circle of radius $R$ centered around the current position $\pos(\tvar)$, where $R=3m$ in this simulation (shown in Figure \ref{fig:init_setup2}).
For a camera, the sensing region $\sense_\tvar$ is determined by a triangular region with solid angle $F$ (also called the field-of-view) and apex at the current vehicle heading, and a maximum extent of $R$. We use $F = \pi/3$ and $R = 20m$ for our simulations (shown in Figure \ref{fig:init_setup3}).
However, part of the regions of $\sense_\tvar$ can be occluded by the obstacles, as would be the case for any real-world sensors.

Additionally, for each sensor, we demonstrate our approach on two different planners $\planner$: a sampling-based planner and a model-based planner.
For the sampling-based planner, we use a Rapidly-Exploring Random Tree (RRT) \cite{lavalle1998rapidly}, and for the model-based planner, we use a spline-based planner~\cite{walambe2016optimal}. 
Our goal is to safely navigate to the goal position despite the unknown obstacles.

\section{Hamilton-Jacobi Reachability} \label{sec:reachability}
Our framework is based on Hamilton Jacobi (HJ) reachability analysis \cite{Margellos11, mitchell2005time}. This analysis has been successfully applied in a variety of domains, such as aircraft auto-landing and safe multi-vehicle path planning \cite{bansal2017safe, bansal2017hamilton}.
In this work, we will be using reachability analysis to compute a backward reachable set (BRS) $\brs(\tdummy)$ given a set of unsafe states $\targetset$.
Intuitively, $\brs(\tdummy)$ is the set of states such that the system trajectories that start from this set can enter $\targetset$ within a time horizon of $\tdummy$ for some disturbance despite the best control efforts.
In contrast, for any trajectory that starts from $\brs^{c}(\tdummy)$, there exists a control such that the system trajectory will \textit{never} enter $\targetset$, despite the worst-case disturbance.
Here, $\brs^{c}(\tdummy)$ represents the complement of the set $\brs(\tdummy)$.

The computation of the BRS can be formulated as a differential game between the control and disturbance, which can be solved using the principle of dynamic programming.
The cost functional corresponding to this differential game is given by:
\begin{equation}
    \label{eqn:warm_cost}
    \costfunctional(\state, \tdummy, \ctrl(\cdot), \dstb(\cdot)) = \inf_{\tdummyy \in [\tdummy,0]} \targetfunc(\trajstandard(\tdummyy)),
\end{equation}
where $\trajstandard(\tdummyy)$ represents the system state at time $\tdummyy$ starting from state $\state$ at time $\tdummy$ and applying control $\ctrl(\cdot)$ with disturbance $\dstb(\cdot)$. 
In \eqref{eqn:warm_cost}, the function $\targetfunc(\state)$ is the implicit surface function representing the unsafe set $\targetset = \{\state: \targetfunc(\state) \le 0\}$. 
Intuitively, $\costfunctional$ keeps track of whether the system trajectory even entered the unsafe set during the time horizon $[\tdummy,0]$, and if so, the cost corresponding to that trajectory is negative. 

The value function corresponding to the cost functional in \eqref{eqn:warm_cost} is given by:
\begin{equation}
    \label{eqn:val_func}
\vfunc(\tdummy, \state) = \min_{\dfunc \in \Gamma[\ctrl]} \max_{\cfunc} \costfunctional(\state, \tdummy, \ctrl(\cdot), \dstb(\cdot)),
\end{equation}
where $\Gamma$ represents the set of non-anticipative strategies \cite{mitchell2005time}.
If the value function is negative for a given state, then starting from this state the system cannot avoid entering into the unsafe set eventually.
Thus, the value function in \eqref{eqn:val_func} keeps track of all unsafe trajectories of the system, which in turn can be used to compute the safe trajectories for the system.
For further details on this formulation, we refer the interested readers to \cite{mitchell2005time, bansal2017hamilton}.

The value function in \eqref{eqn:val_func} can be obtained using dynamic programming, which yields a Hamilton Jacobi-Isaacs Variational Inequality (HJI-VI) \cite{Fisac15, Margellos11}.
Ultimately, a BRS can be computed by solving the following final value HJI-VI: 
\begin{equation}
\label{eq:HJIVI_BRS}
\begin{aligned}
\min \{&D_{\tdummy} \vfunc(\tdummy, \state) + \ham(\tdummy, \state, \nabla \vfunc(\tdummy, \state)), \targetfunc(\state) - \vfunc(\tdummy, \state) \} = 0 \\
&\vfunc(0, \state) = \targetfunc(\state), \quad \tdummy \le 0.
\end{aligned}
\end{equation}
Here, $D_{\tdummy} \vfunc(\tdummy, \state)$ and $\nabla \vfunc(\tdummy, \state)$ denote the time and space derivatives of the value function $\vfunc(\tdummy,\state)$ respectively.
The Hamiltonian, $\ham(\tdummy, \state, \nabla \vfunc(\tdummy, \state))$, encodes the role of system dynamics, control, and disturbance, and is given by
\begin{equation}
\label{eq:HJIVI_ham}
\ham(\tdummy, \state, \nabla \vfunc(\tdummy, \state)) = \max_{\ctrl \in \cset_i} \min_{\dstb \in \dset} \nabla \vfunc(\tdummy, \state) \cdot \dyn(\state, \ctrl, \dstb).
\end{equation}

Once the value function $\vfunc(\tdummy, \state)$ is computed, the BRS, and consequently, the set of safe states are given by
\begin{align}
\brs(\tdummy) = \{\state: \vfunc(\tdummy, \state) \le 0\}, \label{eq:implicitValfuncs} \\
\brsSafe(\tdummy) = \brs^{c}(\tdummy) = \{\state: \vfunc(\tdummy, \state) > 0\} \label{eq:safe_set}. 
\end{align}
HJI reachability also provides the optimal control to keep the system in the safe set and is given by 
\begin{equation}
\label{eq:opt_ctrl}
\ctrl^{*}(\tdummy,\state) =  \arg \max_{\ctrl \in \cset} \min_{\dstb \in \dset} \nabla \vfunc(\tdummy, \state) \cdot \dyn(\state, \ctrl, \dstb).
\end{equation}
In fact, the system can safely apply any control as long as it is not at the boundary of the unsafe region. If the system reaches the boundary of $\brs(\tdummy)$, the control in \eqref{eq:opt_ctrl} steers the system away from the unsafe states. This least restrictive controller provided by HJI reachability is also the basis for ensuring safety for any planner in a least restrictive fashion in our framework.

\section{Our Approach}
We propose an HJ-reachability-based framework to ensure safety in an \textit{a priori} unknown environment.
Our framework also uses a novel, real-time computation of a conservative approximation of the safe set based on new observations of the environment as the vehicle is navigating. 
We first describe our framework, and then present a real-time algorithm to update the safe set.

\subsection{Ensuring safety in unknown environments}
Our framework treats the unsensed environment at any given time as an obstacle. 
The unsensed environment along with the sensed obstacles are used to compute a safe region for the vehicle using HJI reachability. 
This ensures that the vehicle never enters the unknown and potentially unsafe part of the environment, despite the worst case disturbance.

Let $\freeRegion_{\tvar}$ denote the sensed obstacle-free region of the environment at any time $\tvar$.
Given the initial obstacle-free region $\freeRegion_0 = \initSafe$, we compute the corresponding safe set $\brsSafe_0$ by solving the HJI-VI in \eqref{eq:HJIVI_BRS}, assuming everything outside $\freeRegion_0$ is an obstacle.
For this computation, the initial value function in \eqref{eq:HJIVI_BRS} is given by $\vfunc_0(0, \state) := \targetfunc_0(\state)$, where $\targetfunc_0(\state)$ is positive inside $\freeRegion_0$ and negative outside. 
One such function is given by the signed distance to $\freeRegion^{c}_0$.
Starting from $\targetfunc_0(\state)$, the HJI-VI is solved to obtain the value function $\vfunc_0(\state) := \lim_{\tau \to -\infty} \vfunc_0(\tau, \state)$.
Here, $\tau$ is the dummy computation variable in \eqref{eq:HJIVI_BRS}.
$\vfunc_0(\state)$ is then used to compute the safe region $\brsSafe_0$ (see \eqref{eq:safe_set}).
As long as the vehicle is inside $\brsSafe_0$, a controller exists to ensure that it does not collide with the known or unknown obstacles.

We next execute a controller on the system for the time horizon $\tvar \in [0, \horizon]$ as per the following control law:
\begin{equation} \label{eqn:safe_control_law_0}
\ctrl(\tvar)=
    \begin{cases}
      \planner(\state(\tvar), \state^{*}, \env),& \text{if}~\state(\tvar) \in \brsSafe_{\tvar}\\
      \ctrl^{*}(\tvar,\state(\tvar)),& \text{otherwise}
    \end{cases}
\end{equation}
where $\ctrl^{*}(\tvar,\state(\tvar))$ is the optimal safe controller corresponding to $\brsSafe_{\tvar}$ and is given by \eqref{eq:opt_ctrl}.
Also, until the safe set is updated, we use the last computed safe set for finding the optimal safe controller, i.e., $\brsSafe_{\tvar} = \brsSafe_0~\forall \tvar \in [0, \horizon]$.
The control mechanism in \eqref{eqn:safe_control_law_0} is least restrictive in the sense that it lets the planner execute the desirable control on the system, except when the system is at the risk of violating safety.
Note that the control horizon $\horizon$ in our framework can be arbitrarily chosen by the system designer while still ensuring safety.

While the system is executing the control law in \eqref{eqn:safe_control_law_0}, it will obtain new sensor measurements $\sense_{\tvar}$ at each time $\tvar$, which is used to obtain $\freeRegion_{\tvar}$, the free space sensed at that time.
If the sensor is completely occluded by an obstacle at any time, the corresponding free space is an empty set.
Thus, the overall known free space at time $\tvar$ is given by:
\begin{equation} \label{eqn:map}
\map_{\tvar} = \bigcup_{s \in [0, \tvar]} \freeRegion_{\tau}.
\end{equation}
At the end of the control time horizon, we compute another safe region $\brsSafe_{\horizon}$ assuming everything outside $\map_{\horizon}$ is an obstacle. 
This safe region is obtained by solving HJI-VI until convergence.
We then execute a control law similar to in \eqref{eqn:safe_control_law_0}, except that the safety controller intervenes only when the system is at the boundary of $\brsSafe_\horizon$.
The entire procedure is repeated until the system reaches the goal state.

Since the safety controller does not allow the system trajectory to leave the known free space, the proposed framework is guaranteed to avoid collision at all times. 
However, the safe set can be rather conservative especially early on when most of the environment is still unexplored, which is a trade-off we make to ensure safety against all unexpected obstacles.
If additional information about the obstacles in the environment is known, it can be incorporated and will only reduce the conservativeness of the safe set. 

Note that the safe set does not necessarily need to be updated every $\horizon$ seconds. It can be updated faster, slower or at the same rate as the control horizon. 
Essentially one can use the most recent safe set in the control law in \eqref{eqn:safe_control_law_0}, and still ensure safety at all times. 
This is because the safe set at any time $\tvar_1$ is only smaller than the safe set at time $\tvar_2$ when $\tvar_1 < \tvar_2$.
However, the safe set should be updated as quickly as possible to minimize interference with the planner.

\subsection{Efficient update of the BRS}
Our framework requires the computation of a safe set in real-time as the vehicle is navigating through the environment. 
In general, this is challenging due to the exponentially scaling computational complexity of HJI reachability with respect to the state dimension~\cite{bansal2017hamilton}. To mitigate some of the computational challenges, we introduce two novel approaches to computing the BRS: warm-starting and local value function updates. 

\subsubsection{\textbf{Warm-start approach}} \label{sec:warm_start}
At any given time, the vehicle senses only a small additional part of the environment. 
Consequently, the free space map $\map$ only changes by a small amount in a small time horizon.
Intuitively, this should only cause a small change in the safe region. 
We leverage this intuition to propose a novel, faster way to update the reachable set. 
For brevity, we explain our approach assuming that the safe set is updated every $\horizon$ seconds, but the same results hold when the safe set is updated at a non-fixed rate. 

Given the last computed safe set at time $\tvar_{\text{last}}$, and the maps at $\tvar_{\text{last}}$ and the current time $\tvar$, we ``warm-start'' the value function in \eqref{eq:HJIVI_BRS} for the BRS computation at time $\tvar$ as follows:
\begin{equation} \label{eqn:warm_start}
\vfunc_t(0, \state) = 
    \begin{cases}
      \targetfunc_t(\state), & \text{if}~\state \in \map_\tvar \cap \map^{c}_{\tvar_{\text{last}}}\\
      \vfunc_{\tvar_{\text{last}}}(\state), & \text{otherwise}
    \end{cases}
\end{equation}
where $\targetfunc_\tvar(\state)$ as before is defined such that it is positive inside $\map_\tvar$ and negative outside. 
Intuitively, instead of initializing the value function with $\targetfunc_\tvar(\state)$ everywhere in the state space, \eqref{eqn:warm_start} initializes it with the last computed value function for the states where no new information has been obtained since the last computation, and with $\targetfunc_\tvar(\state)$ only at the states which were previously assumed to be occupied but are actually obstacle-free. 
This leads to a much faster computation of BRS because the value function needs to be updated only for a much smaller number of states that are newly found out to be free.
At all the other states, the value function is already almost accurate and only small refinements are required.
Interestingly, this procedure also maintains the conservativeness of the safe region, which is sufficient to ensure collision avoidance at all times.
\begin{lemma} \label{lemma:warm_start}
Let $\tilde{\vfunc}_{\tvar}(\tdummy, \state)$ be the solution of the following warm-started HJI-VI:
\begin{equation*}
\min \{D_{\tdummy} \tilde{\vfunc}_{\tvar}(\tdummy, \state) + \ham(\tdummy, \state, \nabla \tilde{\vfunc}_{\tvar}(\tdummy, \state)), \targetfunc_{\tvar}(\state) - \tilde{\vfunc}_{\tvar}(\tdummy, \state) \} = 0,
\end{equation*}
where $\tilde{\vfunc}_{\tvar}(0, \state)$ is defined as in \eqref{eqn:warm_start}.
Let $\vfunc_{\tvar}(\tdummy, \state)$ be the solution of the HJI-VI in \eqref{eq:HJIVI_BRS} with $V_t(0,x) = l_t(x)$.
Then $\tilde{\vfunc}_{\tvar}(\tdummy, \state) \le {\vfunc}_{\tvar}(\tdummy, \state)$ for all $\tdummy \le 0$. In particular, $\tilde{\brs}_{\tvar}(-\infty) \supseteq {\brs}_{\tvar}(-\infty)$ and $\tilde{\brsSafe}_{\tvar}(-\infty) \subseteq {\brsSafe}_{\tvar}(-\infty)$.

\begin{proof}
Since $\vfunc_{\tvar_{\text{last}}}(\state)$ represents the converged value function at time $\tvar_{\text{last}}$, we have $\vfunc_{\tvar_{\text{last}}}(\state) \le \targetfunc_{\tvar}(\state)~\forall \state \in \R^{n}$. 
Therefore, from \eqref{eqn:warm_start}, $\tilde{\vfunc}_{\tvar}(0, \state) \le \targetfunc_{\tvar}(\state)~\forall \state \in \R^{n}$. Now we are ready to prove our claim.

For $\tdummy \le 0$, by dynamic programming (see \cite{bansal2017hamilton} for details) we know that:
\begin{equation*}
    \label{eq:vl}
    \begin{aligned}
    \tilde{\vfunc}_{\tvar}(\tdummy, \state(\tdummy)) &= \max_\ctrl \min_\dstb \min \Big\{
     \inf_{\tdummyy \in [\tdummy,0)}\targetfunc_{\tvar}(\state(\tdummyy)), \tilde{\vfunc}_{\tvar}(0, \state(0))\Big\},\\
    &\leq \max_\ctrl \min_\dstb \min \Big\{
     \inf_{\tdummyy \in [\tdummy,0)}\targetfunc_{\tvar}(\state(\tdummyy)), \targetfunc_{\tvar}(\state(0))\Big\},\\
    &= {\vfunc}_{\tvar}(\tdummy, \state(\tdummy))
    \end{aligned}
\end{equation*}
where the second inequality follows from the fact that $\tilde{\vfunc}_{\tvar}(0, \state) \le \targetfunc_{\tvar}(\state)~\forall \state \in \R^{n}$.
Thus, from \eqref{eq:implicitValfuncs}, we have that $\tilde{\brs}_{\tvar}(\tdummy) \supseteq {\brs}_{\tvar}(\tdummy)$ for all $\tdummy \le 0$.
In particular, this implies that $\tilde{\brs}_{\tvar}(-\infty) \supseteq {\brs}_{\tvar}(-\infty)$. 
Finally, \eqref{eq:safe_set} implies that $\tilde{\brsSafe}_{\tvar}(-\infty) \subseteq {\brsSafe}_{\tvar}(-\infty)$ since the safe set is the complement of the BRS.
\end{proof}
\end{lemma}
Intuitively, Lemma \ref{lemma:warm_start} states that the safe set obtained by the warm-start approach is an under-approximation of the actual safe set obtained by solving full HJI-VI. 
Thus, it can be used to ensure safety for the vehicle while being computationally efficient. 
In practice, for the sensors and navigation problems in this paper, the amount of conservatism incurred by warm-starting is very small, as we demonstrate in Section \ref{sec:simulations}.
Our overall approach with warm-starting to update the safe set is summarized in Algorithm~\ref{algo:warm_start}.
\begin{algorithm}[tb]
	\DontPrintSemicolon
	\caption{Safe navigation using HJ reachability}
	\label{algo:warm_start}
	$\state_0, \state^{*}$: Start and the goal states\;
	$\freeRegion_0 \longleftarrow \initSafe$: The initial obstacle-free region\;
	$\env$: The unknown environment\; 
	$\planner(\cdot, \state^{*}, \env)$: The planner for the vehicle\; 
	$\horizon$: The control horizon\;
	$\brsSafe_0$: The initial safe region obtained by solving HJI-VI in \eqref{eq:HJIVI_BRS}\;
	$\brsSafe_{\text{last}} \longleftarrow \brsSafe_0$; $\brs_{\text{last}} \longleftarrow \brsSafe^{c}_0$; $\tvar_{\text{last}} \longleftarrow 0$: The last computed safe set, BRS, and the corresponding time\;
	\While{\text{the vehicle is not at the goal}}{	
		Obtain the current sensor observation $\sense_\tvar$ and free space $\freeRegion_\tvar$\;
		Apply the least restrictive control $\ctrl(\tvar)$ given by \eqref{eqn:safe_control_law_0}\;
		\For{\textbf{every} $\horizon$ seconds}{
			Obtain the current map $\map_\tvar$ of the environment using \eqref{eqn:map}\;
			Warm-start the BRS computation using $\brs_{\text{last}}$, $\map$, and \eqref{eqn:warm_start}\; 
			Obtain the new safe region, $\brsSafe_\tvar$, by solving HJI-VI with the warm-started value function\;
			$\brsSafe_{\text{last}} \longleftarrow \brsSafe_\tvar$; $\brs_{\text{last}} \longleftarrow \brsSafe^{c}_\tvar$; $\tvar_{\text{last}} \longleftarrow \tvar$\;
		}
		}
\end{algorithm}
We start with the initial known free space $\initSafe$ and compute the initial safe set $\brsSafe_0$ using HJI-VI (Line 6).
The value function for this computation is initialized by the signed distance to $\initSafe$. 
We also maintain the last computed BRS $\brs_{\text{last}}$, the safe set $\brsSafe_{\text{last}}$, and the corresponding time $\tvar_{\text{last}}$ (Line 7).
At every state, the vehicle obtains the current sensor observation and extracts the sensed free space (Line 9 and 10).
Next, a control command is applied to the vehicle (Line 11). 
If the vehicle is inside $\brsSafe_{\text{last}}$, the planner is used to obtain the control command; otherwise, the safety controller is applied. 
Every $\horizon$ seconds, the safe set and controller are updated based on the free space sensed by the vehicle so far using HJI-VI (Line 15).
The value function for this computation is warm-started with $\vfunc_{\text{last}}$ except at the states which are discovered to be obstacle free since $\tvar_{\text{last}}$ as described in \eqref{eqn:warm_start} (Line 14).
The whole procedure is repeated until the vehicle reaches its goal.

\subsubsection{\textbf{Local update of the BRS}} \label{sec:localQ}
In the last section, we discussed how warm-starting the value function computation might lead to a faster convergence of the value function; however, the value function is still computed over the entire state space.
In this section, we present a more practical algorithm that leverages the advantages of warm-starting by computing and updating the value function only locally at the states for which new information has been obtained since the last value function computation.
Our safety framework is still same as what described in Algorithm \ref{algo:warm_start}---only the computational procedure for the safe set computation (Line 15 in Algorithm \ref{algo:warm_start}) is being modified to update the value function locally. 
We outline this procedure in Algorithm \ref{algo:localQ}.
\begin{algorithm}[tb]
	\DontPrintSemicolon
	\caption{Local update of the BRS}
	\label{algo:localQ}
	$\queue \longleftarrow \map_\tvar \cap \map^{c}_{\tvar_{\text{last}}}$: Initialize list of states for which the value function should be updated\;
	$\queue \longleftarrow \queue \cup \neighbor(\queue)$: Add neighboring states to $\queue$\;
	Warm-start the value for states in $\queue$, $\vfunc_{\tvar}(0, \queue)$, using \eqref{eqn:warm_start}\;
	$\vfunc_{\text{old}} \longleftarrow \vfunc_{\tvar}(0, \queue)$: The last computed value function for states in $\queue$\; 
	\While{$\queue$ \text{is not empty}}{	
	    $\vfunc_{\text{updated}} \longleftarrow$ Update the value function $\vfunc_{\text{old}}$ for a time step $\timestep$\;
	    $\Delta\vfunc = \|\vfunc_{\text{updated}} - \vfunc_{\text{old}}\|$: Change in the value function\;
		$\queue_{\text{remove}} \longleftarrow \{\state \in \queue: \Delta\vfunc = 0\}$: States with unchanged value\;
		$\queue \longleftarrow \queue - \queue_{\text{remove}}$: Remove states with unchanged value\;
		$\queue \longleftarrow \queue \cup \neighbor(\queue)$: Add neighboring states to $\queue$\;
		$\vfunc_{\text{old}} \longleftarrow \vfunc_{\text{updated}}$\;
		}
\end{algorithm}
\setlength{\textfloatsep}{0pt}

In Algorithm \ref{algo:localQ}, we maintain a list of states $\queue$ at which the value function needs to be updated in light of the new environment observations. 
$\queue$ is initialized to be the set of states that are newly discovered to be free since $\tvar_{\text{last}}$, i.e., $\queue = \map_\tvar \cap \map^{c}_{\tvar_{\text{last}}}$ (Line 1).
Since the change in the value of the states in $\queue$ (compared to $\vfunc_{\tvar_{\text{last}}}(\state)$) would also cause a change in the value of the neighboring states, $\neighbor(\queue)$, we also add them to $\queue$ (Line 2).
Thus, $\queue = \queue \cup \neighbor(\queue)$.
Typically, the value function in HJI-VI is computed by discretizing the state-space into a grid and solving the VI over that grid.
In such cases, the spatial derivative of the value function (required to compute the Hamiltonian in the HJI-VI in \eqref{eq:HJIVI_BRS}) is computed numerically using the neighboring grid points. 
This spatial derivative is precisely responsible for the propagation of the change in the value function at a state to its neighboring states. 
In such cases, $\neighbor(\queue)$ might represent the neighboring grid points used to compute the spatial derivative of the value function for the states in $\queue$; however, other neighboring criteria can be used.

Once the neighbors are added to $\queue$, the value for all the states in $\queue$ is initialized as per \eqref{eqn:warm_start} (Line 3), and their value is updated using HJI-VI in \eqref{eq:HJIVI_BRS} for some time step $\timestep$ (Line 6). 
This computation is much faster than classical HJI-VI computation since it is typically performed for many fewer states.
Next, we remove all those states from $\queue$ whose value function hasn't changed significantly over this $\timestep$ (Line 8 and 9), as these states won't cause any change in the value function for any other state.
The neighbors of the remaining states are next added to $\queue$ (Line 10) and the entire procedure is repeated until the value function is converged for all states.
Note that Algorithm \ref{algo:localQ} still maintains the conservatism of the safe set since it is just a different computational procedure for computing the warm-started value function, which is still used within the safety framework in Algorithm \ref{algo:warm_start}.

\section{Simulations} \label{sec:simulations}
\subsection{Running example revisited}
\begin{figure} [h!]
    \centering
    \includegraphics[width=\columnwidth]{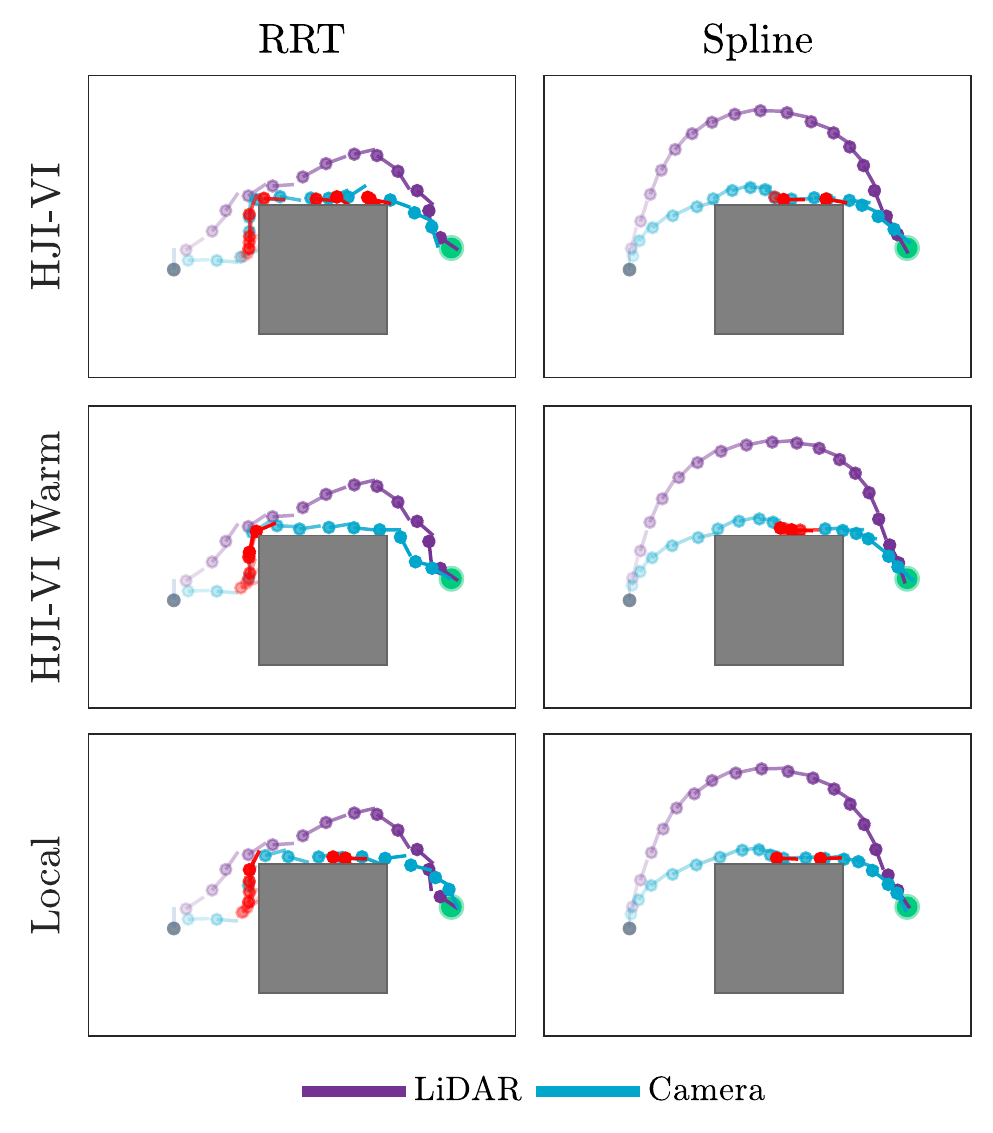}
    \caption{The vehicle trajectories for the problem setting in Figure \ref{fig:initial_setup_running_example} for both planners (RRT and Spline planners) and both sensors (LiDAR and Camera sensors) with the safety controller computed from each of the three candidate safety approaches.
    The proposed framework is able to safely navigate the vehicle to the goal in all cases.
    When the planner makes unsafe decisions, the safety controller intervenes (the states marked in red) to ensure safety.}
    \label{fig:trajectory_results}
\end{figure}

We now return to our running example and demonstrate the proposed approach in simulation (described in Section \ref{sec:running_example}). 
We implement our safety framework with three different methods to update the BRS: using the full HJI-VI, the warm-start approach (Section \ref{sec:warm_start}), and the local update approach (Section \ref{sec:localQ}).
The corresponding system trajectories for different planners and sensors for all the three methods are shown in Figure \ref{fig:trajectory_results}.
For all combinations of planners and sensors, the proposed framework is able to safely navigate the vehicle to its goal position despite the external disturbances and no \textit{a priori} knowledge of the obstacle (none of the trajectories go through the obstacle).
As the vehicle navigates through the environment, the planner makes optimistic decisions at several states that might lead to a collision; however, the safety controller intervenes to ensure safety. States where the safety controller is applied are marked in red. 
Note that the safety controller intervenes more frequently for the camera sensor as compared to the LiDAR. 
This is because the field-of-view (FoV) of a camera is typically much narrower than a LiDAR (which senses the obstacles in all directions within a range).
Given this limited FoV, the safety controller needs to account for a much larger unexplored environment, which in turn leads to more cautious control.

We compare the computation time required for each of the three methods to compute the BRS for the camera and LiDAR sensors in Table \ref{tab:simulated_results}.
All computations were done on a MATLAB implementation on a desktop computer with a Core i7 5820K processor using the Level Set toolbox \cite{mitchell2004toolbox}.
As expected, across all scenarios, warm-starting the value function for the BRS computation leads to a significant improvement in computation time compared to full HJI-VI; however, the computation time might still not be practical for most real-world applications.
Only locally updating the value function in addition to warm-starting leads to a significant further improvement in the computation time, and the BRS is updated in approximately 1s on average for all sensors and planners. 
This improvement is impressive considering that the computation was done in MATLAB without any parallelization which is known to decrease the computation time by a factor of 100 \cite{chen2017provably}.

Lemma \ref{lemma:warm_start} indicates that the safe set obtained by warm-starting the value function is conservative compared to the one obtained by full HJI-VI.
Therefore, we also compare the percentage volume of the states at which the safe set is conservative. 
This over-conservative volume is typically limited to 0.5\% which indicates that the warm-starting approach is able to approximate the true value function quite well. 
%
{\renewcommand{\arraystretch}{1.5}
\begin{table}[t!]
\centering
\begin{subtable}{1\columnwidth}
\begin{tabular}{ |l|l|l|l|l| }
\hline
\rowcolor{turquoise}
\multicolumn{5}{ |c| }{\textbf{Simulated Camera Results}} \\
\hline
\rowcolor{lightTurquoise}
Metric & Planner & HJI-VI & Warm & Local \\ \hline
\multirow{2}{*}{Average Compute Time (s)} & RRT     & 45.688 & 26.290 & 0.596 \\ \cline{2-5}
                                          & Spline  & 51.723 & 12.489 & 0.898 \\ \hline
\multirow{2}{*}{\% Over-conservative States} & RRT      & 0.0 & 1.112 & 0.517 \\ \cline{2-5}
                                             & Spline   & 0.0 & 0.474 & 0.506 \\
\hline
\end{tabular}
\end{subtable}

\vspace{2mm}
%
\begin{subtable}{1\columnwidth}
\begin{tabular}{ |l|l|l|l|l| }
\hline
\rowcolor{purple}
\multicolumn{5}{ |c| }{\textbf{Simulated LiDAR Results}} \\
\hline
\rowcolor{lightPurple}
Metric & Planner & HJI-VI & Warm & Local \\ \hline
\multirow{2}{*}{Average Compute Time (s)} & RRT     & 21.145    & ~6.075         & 1.108 \\ \cline{2-5}
                                      & Spline      & 25.318    & ~3.789     & 1.158 \\ \hline
\multirow{2}{*}{\% Over-conservative States} & RRT & 0.0 & ~0.032 & 0.290 \\ \cline{2-5}
                                        & Spline & 0.0 & ~0.024 & 0.240 \\
\hline
\end{tabular}
\end{subtable}
\caption{Numerical comparison of average compute time and relative volume of over-conservative states for each planner and sensor across different BRS update methods.
Local updates compute an almost exact BRS in $\approx$1 second, and are significantly faster than both HJI-VI and warm-start.} 
\label{tab:simulated_results}
\end{table}}

\begin{figure}
    \centering
    \begin{subfigure}{0.45\columnwidth}
        \centering
        \includegraphics[width=\columnwidth]{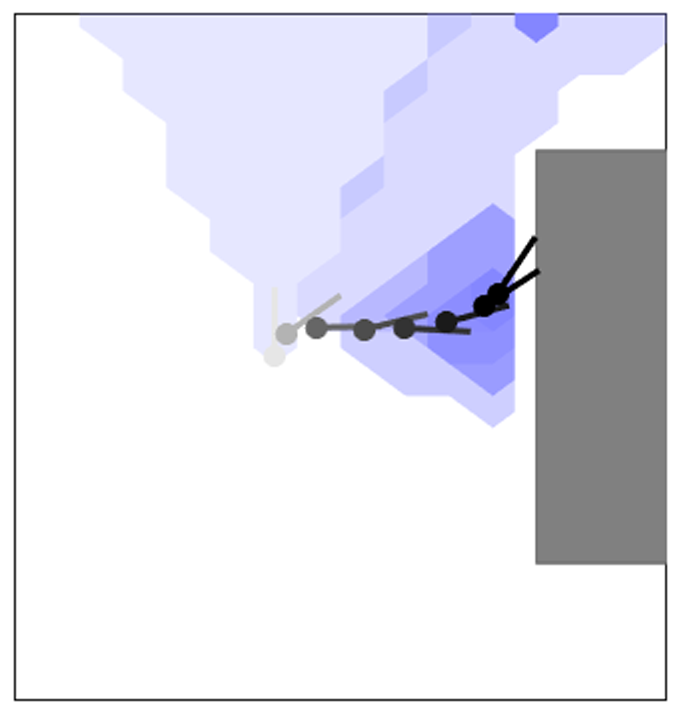}
        \subcaption{}
        \label{fig:sim1_sub1}
    \end{subfigure}
    \begin{subfigure}{0.45\columnwidth}
        \centering
        \includegraphics[width=\columnwidth]{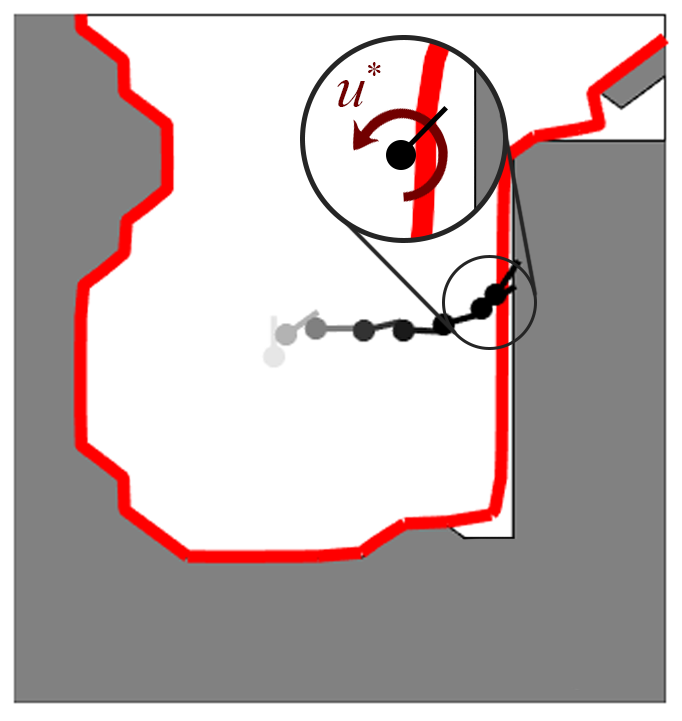}
        \subcaption{}
        \label{fig:sim1_sub2}
    \end{subfigure}
    \caption{(a) The sensed region by the vehicle at different states in time for the camera sensor. (b) The overall free space sensed by the vehicle and the corresponding safe set (interior of the red boundary). Since the vehicle is at the boundary of the safe set, the safety controller $\ctrl^{*}$ is applied to steer the robot inside the safe set and ensure collision avoidance.}
    \label{fig:safety_snapshot}
\end{figure}

Finally, we take a closer look at how the safe control comes into play when the system is operating with a range-limited sensor. Figure \ref{fig:sim1_sub1} showcases a Dubins' car with a camera sensor and an RRT planner, where the current robot state is shown in black, the corresponding sensed region is in dark blue, and the trajectory and corresponding sensed regions are shown in grey and light blue respectively.
Since the camera's FoV is occluded by an obstacle at the current state, it cannot sense the environment past the obstacle.
Figure \ref{fig:sim1_sub2} illustrates the corresponding current belief map of the environment which is the union of the free space sensed by the vehicle so far (shown in white).
Since the current sensed region is contained within the sensed region at the previous state, no new environment information is obtained and hence the BRS is not updated.
The slice of the safe set at the current vehicle heading is shown in Figure \ref{fig:sim1_sub2} (the area within the red boundary).
Since the vehicle is at the boundary of the safe set, the safety controller intervenes and applies a control $\ctrl^{*}$ that leads the system towards the interior of the safe set (the red arrow) to ensure collision avoidance.

\subsection{Safety for a learning-based planner} \label{sec:leanring_planner}
Since the proposed safety framework is planner-agnostic, we can use it to ensure safe navigation even in the presence of a learning-based planner. 
In particular, we use the vision-based planner proposed in \cite{bansal2019combining}, which takes an RGB camera image and the goal position as input, and uses a Convolutional Neural Network-based perception module to produce a desired next state that moves the robot towards its goal while trying to avoid obstacles on its way. 
This desired next state is used by a model-based low-level planner to produce a smooth trajectory from the vehicle's current state to the desired state.
The authors demonstrate that the proposed planner can leverage robot's prior experience to navigate efficiently in novel indoor cluttered environments; however, it still leads to collisions in several real-world scenarios, like when the vehicle needs to go through narrow spaces.
We use the proposed safety framework to ensure both safe and efficient planning in such difficult navigation scenarios.

\begin{figure}
\centering
    \begin{subfigure}{0.46\linewidth}
        \centering
        \includegraphics[width=\columnwidth]{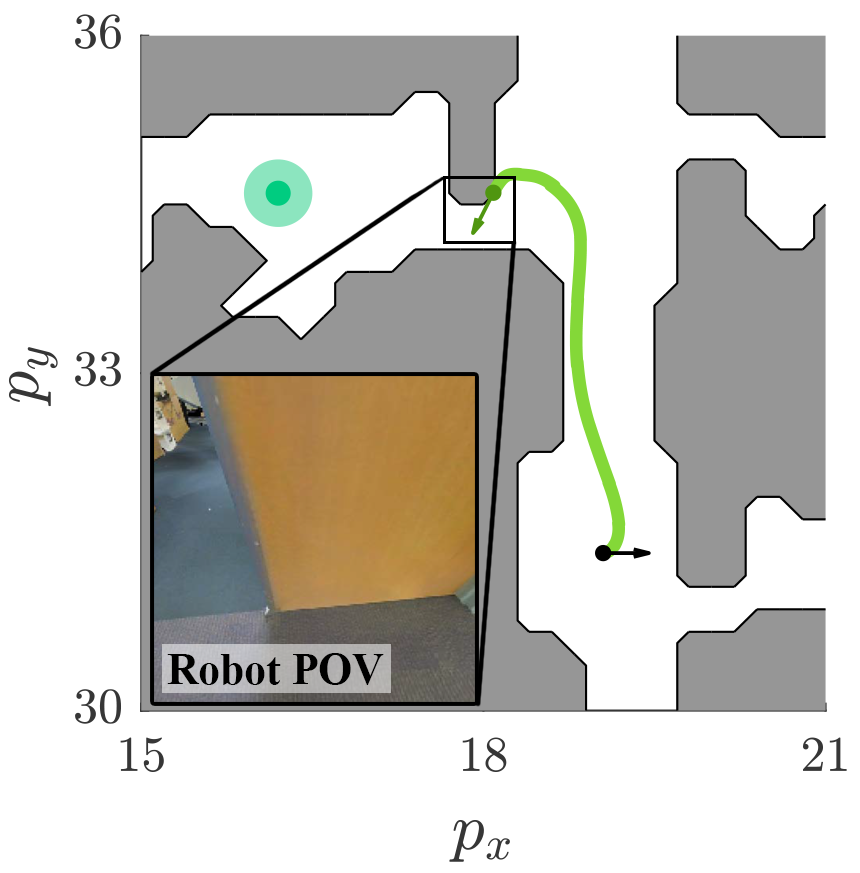}
        \subcaption{}
        \label{fig:sub1}
    \end{subfigure}
    \begin{subfigure}{0.46\linewidth}
        \centering
        \includegraphics[width=\columnwidth]{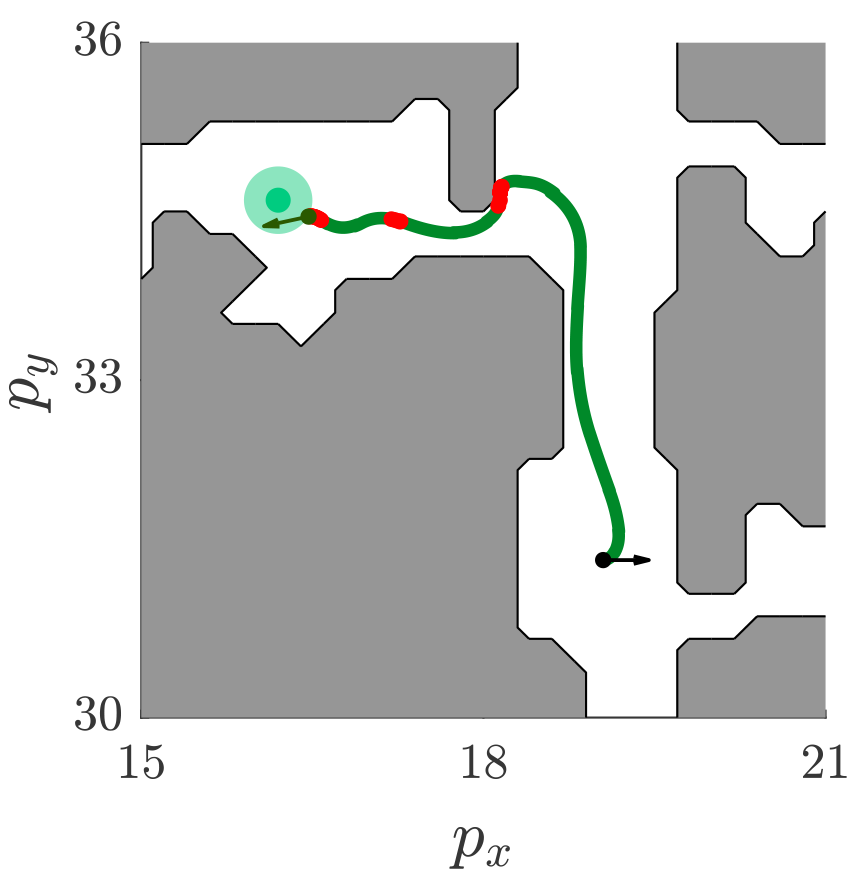}
        \subcaption{}
        \label{fig:sub2}
    \end{subfigure}
    \caption{The proposed framework can be exploited to provide safety guarantees around vision-based planners that incorporate learning in the loop. 
    The vision-based planner plans a path through the doorway. Without safety control (a) this results in collision, however with safety (b) the robot avoids collision and reaches the goal.}
    \label{fig:neural_net_snapshots}
\end{figure}

The task setup for our simulation is shown in Figure \ref{fig:sub1}. 
The robot needs to go through a very narrow hallway, followed by a door into the room to reach its goal (the green circle) starting from the initial state (black arrow). 
At the beginning, the robot has no knowledge about the obstacles (shown in dark grey).
We simulate this scenario using the S3DIS dataset which contains mesh scans of several Stanford buildings \cite{2017arXiv170201105A}. 
By rendering this mesh at any state, we can obtain the image observed by the camera (used by the planner) as well as the occupancy information within the robot's FoV (used for the safety computation).
For the robot dynamics, we use the 4D Dubins' car model:
\begin{equation} \label{eqn:4D_dubins_dyn}
\begin{aligned}
\dot{p}_x = v\cos\phi,\quad \dot{p}_y = v\sin\phi, \quad \dot{v} = a, \quad \dot{\phi} = \omega
\end{aligned}
\end{equation}
where $\pos = (p_x, p_y)$ is the position, $\phi$ is the heading, and $v$ is the speed of the vehicle.
The control is $\ctrl := (a, \omega)$, where $|a| \le 0.4$ is the acceleration and $|\omega| \le 1.1$ is the turn rate.

The trajectory taken by the learning-based planner in the absence of the safety module is shown in Figure \ref{fig:sub1}. 
Even though the vehicle is able to go through the narrow hallway, it collides with the door eventually.
The trajectory taken by the vehicle when the planner is combined with the proposed safety framework is shown in Figure \ref{fig:sub2}. 
When the planner takes an unsafe action near the door, the safety controller intervenes (marked in red) and guides the robot to safely go through the doorway.
We also illustrate the image observed by the robot near the doorway in Figure \ref{fig:sub1}.
Even though most of the robot's vision is blocked by the door, the planner makes a rather optimistic decision of moving forward and leads to a collision.
In contrast, the safety controller makes a conservative decision of rotating in place to explore the environment more before moving forward, and eventually goes through the doorway to reach the goal. The planner-agnostic nature of our framework allows us to provide safety guarantees around learning-based planners as well.

\section{Experiments} \label{sec:experiments}
We test the proposed approach in hardware using a TurtleBot 2 with a mounted stereo RGB camera. For the vehicle state measurement, we use the on-board odometry sensors on the TurtleBot. In our experiment, the vehicle needs to navigate through an unknown cluttered indoor environment to reach its goal (shown in Figure \ref{fig:hardware_front_fig}).
For the BRS computation, we use the dynamics model in \eqref{eqn:4D_dubins_dyn}.
We pre-map the environment using an open-source Simultaneous Localization and Mapping (SLAM) algorithm and the on-board stereo camera. This pre-mapping step is used to avoid the significant delay and inaccuracies in the real-time SLAM map update. However,  the full map is not provided to the robot during deployment. Instead, for the safe set computation at any given time, the current FoV of the camera is projected on the SLAM map and only the information within the FoV is used to update the safe set. This emulates the limited sensor range of the Turtlebot's camera. Regardless, this alludes to one of the important and interesting future research directions of ensuring safety despite sensor noise.

\begin{figure} [t!]
    \centering
    \includegraphics[width=\columnwidth]{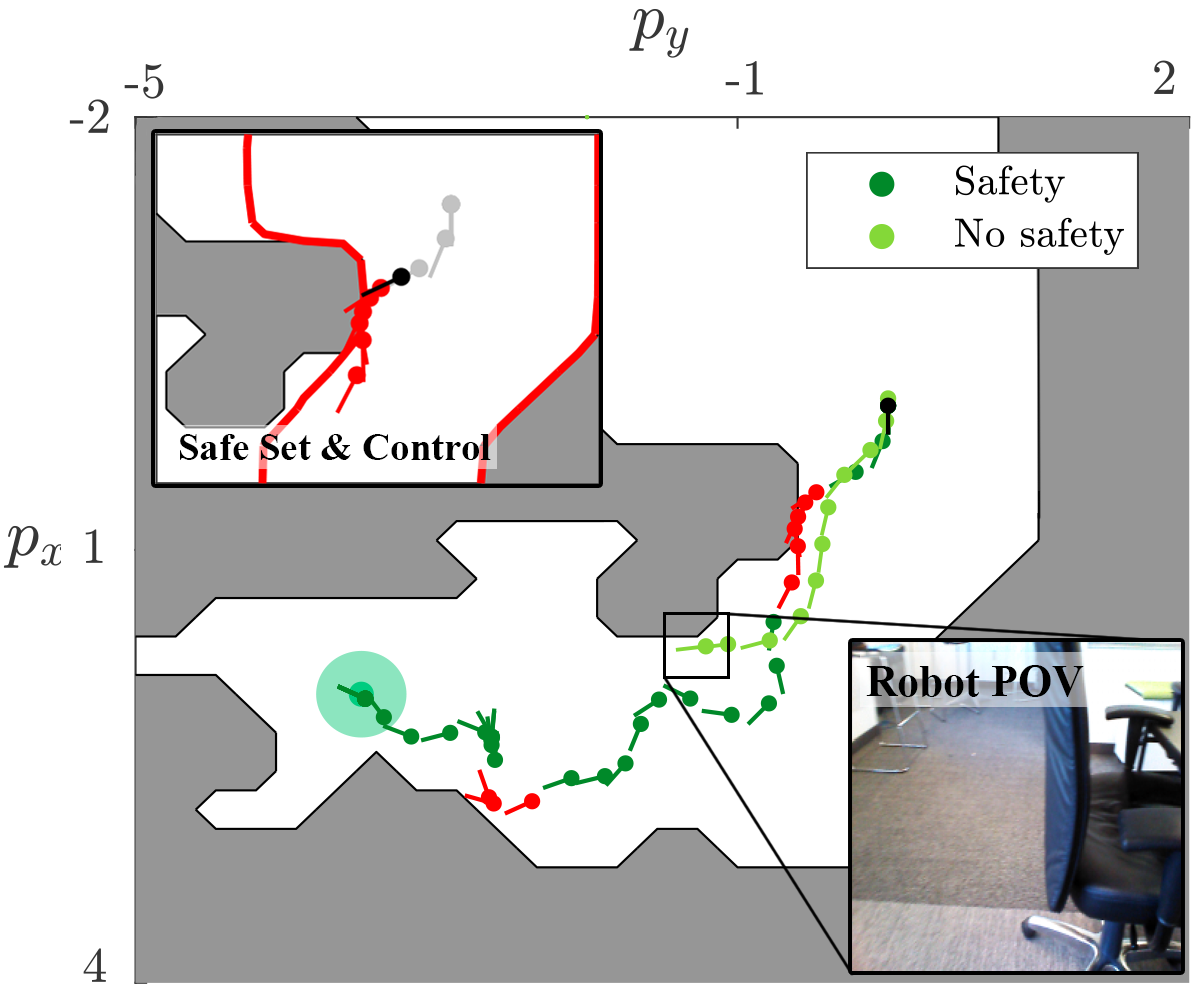}
    \caption{We show the top-view of our experiment setting, and the corresponding system trajectories with and without the proposed safety framework.
    Without the safety framework, the robot collides into the chair.
    In contrast, our safety framework is able to safely navigate the robot to its goal by intervening when the vehicle is too close to the obstacles.}
    \label{fig:experiment_snapshots}
\end{figure}

For  planning, we use the learning-based planner described in Section \ref{sec:leanring_planner} that uses the current RGB image to determine a candidate next state. A top-view of our experiment setting is shown in Figure \ref{fig:experiment_snapshots}.
The vehicle starting position and heading are shown in black, the goal region is shown in green, and the obstacles (unknown to the vehicle beforehand) are shown in grey.
We ran the experiment with and without the safety controller and show the corresponding trajectories in Figure \ref{fig:experiment_snapshots}.
Without the safety controller, the learning-based planner struggles with making sharp turns near the corner, and eventually collides into the obstacle (the chair, in this case).
For context, we also show the RGB observation received by the planner near the corner.
Even though the robot is very close to the chair, the planner makes the unsafe decision of continuing to move forward. 
However, when the learning-based planner is used within the proposed safety framework, the safety controller is able to account for this unsafe situation and safely steer the vehicle away from the obstacle.
We show the corresponding safe set when the vehicle is at the obstacle boundary and the corresponding vehicle trajectory obtained using the safety controller. Afterwards, the planner takes over and steers the vehicle to the goal.

\section{Practical Considerations} \label{sec:practical}
When implementing a safety framework on real systems, there are many practical considerations that should be acknowledged. Below we discuss some of the main practical considerations we encountered during our simulation and hardware experiments when using our proposed framework:
\begin{itemize}
    \item Since the value function is computed over a discretized state space, it might incur some numerical inaccuracies. 
    Using a finer discretization and a higher order approximation for the spatial derivative is often helpful in alleviating these issues; however, the computation time also increases consequently. 
    In our experience, we have found the 3rd or higher order approximations schemes to work pretty well. 
    
    \item Due to the complicated geometry of real-world obstacles, the sensed map at any given time could appear highly irregular. 
    Such irregular maps induce irregular implicit surface functions, which can significantly hamper the value function computation. Thus, it might often be desirable to convert the occupancy map into a regular, well-behaved function, such as a signed distance map, and use that for the value function computation. 
    
    \item Theoretically speaking, the safety controller only needs to be applied when the vehicle is at the boundary of the safe set. However, in practice, due to numerical inaccuracies, it might be desirable to apply the safety controller at a positive level of the value function.  
\end{itemize}

\section{Conclusion And Future Work} \label{sec:conclusion}
Providing safety guarantees for real-world autonomous systems operating  in \textit{a priori} unknown environments is a challenging but important problem. 
In this paper, we propose an HJ reachability-based safety framework for navigation in unknown environments that is applicable to a wide variety of planners and sensors. 
To overcome the computation complexity of classical HJ reachability analysis, we propose a novel, real-time algorithm to update the reachable set as the vehicle traverses the environment.
We demonstrate our approach on multiple sensors and planners, including a learning-based planner, both in simulation and on a hardware testbed.

Several interesting future directions emerge from this work.
First, the proposed framework currently assumes perfect state estimation and sensor measurements. However, these assumptions may not hold in practice, and need to be appropriately accounted for in the framework. Additionally, the current safety guarantees hold for static environments--extensions to dynamic, multi-agent environments is another interesting and valuable research direction.

\section*{Acknowledgments} \label{sec:acknowledge}
This research is supported in part by the DARPA Assured Autonomy program under agreement number FA8750-18-C-0101, by NSF under the CPS Frontier project VeHICaL project (1545126), and by SRC under the CONIX Center, and by Berkeley Deep Drive.

The authors would also like to thank Kene Akametalu, Ellis Ratner, and Anca Dragan for their helpful advice on various technical issues examined in this paper.

\bibliographystyle{plain}
\bibliography{arXiv} 

\end{document}